\documentclass{article}

\usepackage{arxiv}

\usepackage[utf8]{inputenc} 
\usepackage[T1]{fontenc}    
\usepackage{hyperref}       
\usepackage{url}            
\usepackage{booktabs}       
\usepackage{amsfonts}       
\usepackage{nicefrac}       
\usepackage{microtype}      
\usepackage{lipsum}
\usepackage{graphicx}
\usepackage{amsmath}
\graphicspath{ {./images/} }
\usepackage{tikz}
\usepackage{xcolor}
\usepackage{float}
\usepackage{tabularx} 
\usepackage{enumitem}
\usepackage{array}
\usepackage{makecell}
\usepackage{caption}
\usepackage{wrapfig}
\usepackage{amsmath}
\usepackage{amssymb}
\usepackage{mathtools}
\usepackage{amsthm}

\title{Memorization Control in Diffusion Models from Denoising-centric Perspective}

\author{
Thuy Phuong Vu \\
Department of Computer Science\\
Brunel University of London\\
London, United Kingdom\\
\And
Mai Viet Hoang Do \\
Hanoi University of Science and Technology \\
Hanoi, Vietnam \\
\And
Minhhuy Le\\
Faculty of Electrical and Electronic Engineering\\
School of Engineering\\
Phenikaa University\\
Hanoi, Vietnam \\
\And
Dinh-Cuong Hoang \\
Greenwich Vietnam\\
FPT University\\
Hanoi, Vietnam \\
\And
Phan Xuan Tan \textsuperscript{*}\\
College of Engineering\\
Shibaura Institute of Technology\\
Tokyo, Japan\\
}

\begin{document}
\maketitle

\begin{abstract}
Controlling memorization in diffusion models is critical for applications that require generated data to closely match the training distribution. Existing approaches mainly focus on data-centric or model-centric modifications, treating the diffusion model as an isolated predictor. In this paper, we study memorization in diffusion models from a denoising-centric perspective. We show that uniform timestep sampling leads to unequal learning contributions across denoising steps due to differences in signal-to-noise ratio, which biases training toward memorization. To address this, we propose a timestep sampling strategy that explicitly controls where learning occurs along the denoising trajectory. By adjusting the width of the confidence interval, our method provides direct control over the memorization–generalization trade-off. Experiments on image and 1D signal generation tasks demonstrate that shifting learning emphasis toward later denoising steps consistently reduces memorization and improves distributional alignment with training data, validating the generality and effectiveness of our approach.
\end{abstract}

\section{Introduction}
Diffusion models (DMs) \cite{ho2020denoising} have been widely adopted across a broad range of applications, extending beyond image generation and image editing to domain-specific tasks such as data augmentation in scientific and industrial settings \cite{11268708}. Unlike creative image generation, where diversity and novelty are highly desired, many domain-specific applications require generated samples to strictly follow the training data distribution, rendering excessive creativity unnecessary or even harmful. In such scenarios, the ability to control memorization in diffusion models becomes critically important.

Recent studies have established, both empirically and theoretically, that memorization can indeed arise in diffusion models \cite{halder2024memorization, achilli2025memorization}. In particular, \cite{song2025selective} demonstrate that the sampling process of diffusion models can be partitioned into generalization and memorization regimes. During early denoising steps, when noise levels are high, the generated samples exhibit more general, distribution-level characteristics, whereas later steps closer to the final output introduce increasingly specific and detailed structures. Moreover, the authors show that the sampling trajectory naturally drifts into an extrapolation regime. As a consequence, deviations introduced during early denoising steps when compounded by the largely deterministic nature of the denoising process can prevent the generated samples from converging back to the training data distribution.

While prior work has explored controlling memorization through data manipulation, model capacity, architectural design, or loss-based regularization, such approaches largely treat diffusion models as isolated predictors. In contrast, diffusion models fundamentally operate as components within a sequential denoising process, where model predictions are repeatedly applied and accumulated over time. Therefore, adjusting memorization behavior in diffusion models requires modifying not only what the model learns, but also how it learns within the denoising dynamics. Motivated by this perspective, we approach memorization control from a denoising-centric viewpoint, explicitly accounting for the role of the model as an integral part of the denoising process.

In this paper, we make the following contributions. First, we provide a theoretical analysis of diffusion model training through denoising perspective, showing that despite uniform timestep sampling, different denoising steps contribute unequally to the optimization objective. We formally demonstrate that this imbalance leads to biased denoising behavior, which in turn contributes to the emergence of memorization. From this observation, we propose a denoising-centric training strategy that explicitly adjusts the model’s learning behavior. Rather than encouraging uniform learning across both generalization and memorization regimes, we design the model to focus on learning effectively in the more challenging regime, thereby preventing denoising trajectories from drifting into extrapolation regions. Finally, we evaluate our hypothesis on both image generation and 1D data generation tasks, demonstrating that our approach is effective not only for highly creative tasks such as 2D image synthesis, but also for domain-specific scenarios where controlled memorization is required.
\section{Related works}
Diffusion models learn a score function from noisy data and use it during sampling to generate images \cite{ho2020denoising}. Memorization and generalization in DMs exist and the nature of the transition from memorization to generalization are mathematically shown in research of \cite{halder2024memorization, achilli2025memorization}. If the model were to learn the exact score of the training data everywhere in the input space, it would merely reproduce the training samples and fail to generate novel data. Instead, the model underfits only in certain regions of the input space while fitting well in others, a phenomenon termed selective underfitting \cite{song2025selective}, which challenges the view that diffusion models underfit the entire score function. The authors distinguish between a supervision region and an extrapolation region, but do not provide a concrete mechanism to explicitly control the trade-off between these two regions.

To address memorization, \cite{10656527} introduce a framework that adjusts model outputs during sampling to prevent memorization while preserving generation quality. Unlike prior studies focusing on training dynamics or model architecture, it operates at inference time, selectively intervening step-by-step only when memorization is detected, thus minimizing disruption to the standard sampling process. However, since this approach intervenes only at the sampling stage and primarily filters or redirects outputs, it does not address the root causes of memorization, making the mitigation inherently dependent on stochastic sampling behavior.

Recently, many works impress on the importance of data in their memorization. \cite{achilli2025memorization} highlights the critical role of data geometry in memorization, mathematically showing that diffusion models’ tendency to memorize is strongly influenced by the manifold structure of the training data. \cite{bonnaire2025why} attributes the lack of memorization in diffusion models to implicit dynamical regularization during training, with the generalization–memorization timescale separation depending on dataset size. Additionally, \cite{yoon2023diffusion} states that generalization and memorization are mutually exclusive phenomena and by reducing model capacity and complexity of train data as injecting dummy data, it can be prevented and increase generalization. Moreover, memorization is also affected by time embedding method or some training configurations as training epochs, batch size, weight decay \cite{gu2023memorization}. \cite{buchanan2025edgememo} also state that transition from generalization to memorization is governed by the loss landscape and depends on both data complexity and model capacity. They identify a crossover point between them, determined by the intersection of training losses for generalizing and memorizing denoisers. However, \cite{ye2025provable} explains memorization through a provable separation between empirical and ground-truth objectives and mitigates it via model-capacity control rather than loss optimization. While prior work studies memorization from data, model, or training perspectives, it largely overlooks the multi-step denoising nature of diffusion models, in which memorization may emerge as learned biases are repeatedly carried over and strengthened across denoising steps.

Unlike earlier works that attribute memorization to data, training timescales, or loss landscape transitions, \cite{kim2025diffusion} emphasizes the role of data–model interactions during denoising. The authors explain memorization as a consequence of biased early denoising predictions amplified by classifier-free guidance. However, early timesteps receive limited training signal due to weak gradient contributions, raising the question of whether these early predictions are sufficiently informative to meaningfully guide the sampling process. 

Motivated by these observations, we control memorization in diffusion models by adjusting model learning from a denoising-centric perspective, explicitly strengthening the contribution of critical denoising stages during training.

\section{Theoretical analysis of gradient contributions across denoising timesteps}
\begin{figure*}[ht]
    \centering
    \includegraphics[width=\textwidth]{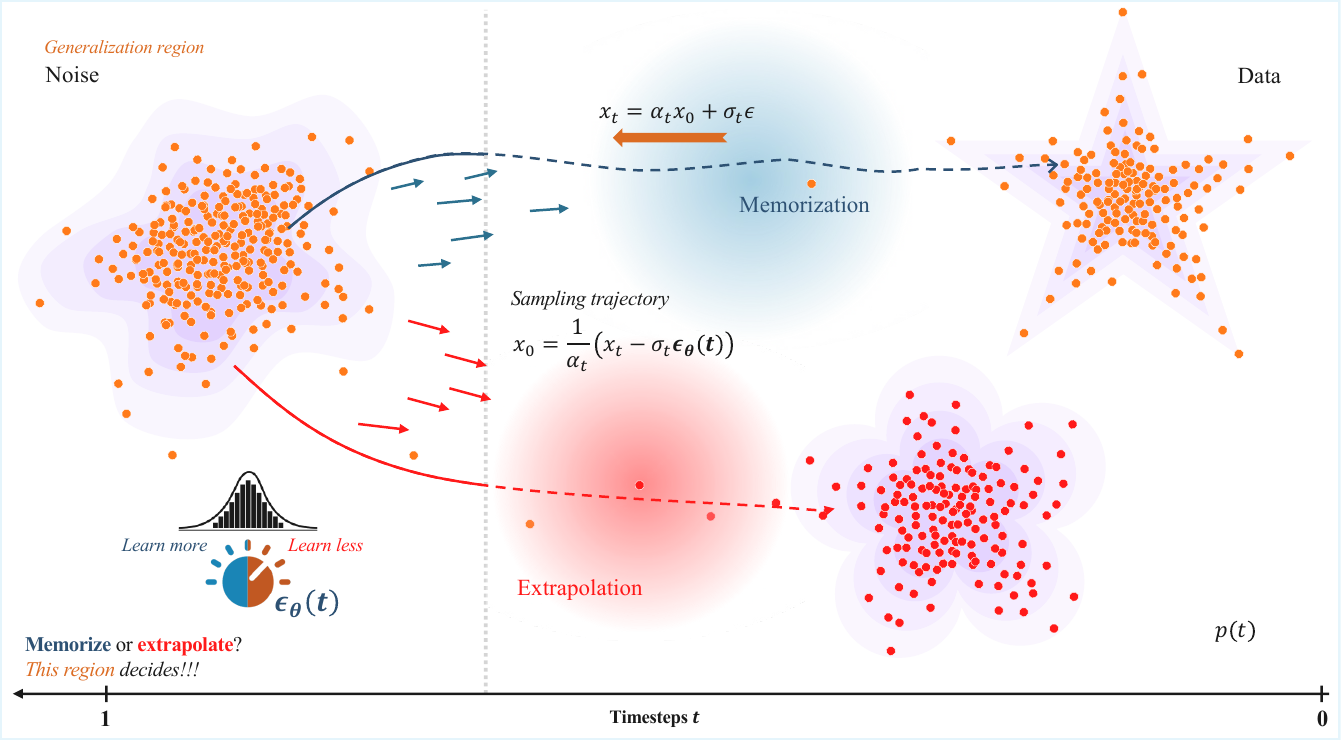}
    \caption{Illustration of the diffusion sampling trajectory, where early denoising steps play a critical role in determining whether the trajectory moves toward memorization or extrapolation. The trajectory is guided by the noise predicted by the denoising model at each timestep. Biases introduced in these early steps can accumulate over time, directing the sample toward the training data manifold or away from it. In this view, the diffusion model acts as a control knob for memorization by shaping the denoising dynamics along the sampling process.}
\end{figure*}
In DDPM training \cite{ho2020denoising}, we sample timesteps as $t \sim \mathcal{U}(\{1, \dots, T\})$ with the purpose of minimizing the loss function $\mathcal{L}(\theta)=\mathbb{E}_{x_0, \epsilon, t \sim p(t)}\left[\left\|\epsilon - \epsilon_\theta(x_t, t) \right\|_2^2\right]$. Superficially, the sampling method makes timesteps contribute equally but equal sampling probability does not imply equal optimization influence and the loss at different $t$ does not have comparable scale or information. In forward process, $x_t = \alpha_t x_0 + \sigma_t \epsilon$ where $\alpha_t = \prod_{s=1}^{t} \sqrt{1 - \beta_s}$ and $\bar{\alpha}_t = \prod_{s=1}^{t} (1 - \beta_s)$ (with $\beta_t \in (0,1), \quad t = 1, \dots, T$) so $\alpha_t \downarrow 0$ and $\sigma \uparrow 1$ as $t$ increases. It immediately implies that with small $t$, the noisy sample is data-nominated $x_t \approx x_0$ while with large $t$, it is noise-nominated $x_t \approx \epsilon$.

As in Eq. \ref{eq:gradient}, the expected gradient magnitude depends on how predictable $\epsilon$ is from $x_t$. This predictability is governed by the signal-to-noise ratio as Eq. \ref{equation:snr}.
Therefore, at small $t$, SNR is high, the model sees strong correlations between $x_t$ and $x_0$, gradients align across samples so this is the memorization regime. In contrast, at large $t$, SNR is low, $x_t$ is nearly pure noise, the denoising task becomes generic and gradients have lower magnitude and less alignment. The expected contribution of timestep $t$ to learning is approximately as Eq. \ref{equation:gradientsnr} with uniform $t$ sampling. 
\begin{equation}
    \nabla_\theta \ell_t
= 2 \left( \epsilon_\theta(x_t, t) - \epsilon \right)
\nabla_\theta \epsilon_\theta(x_t, t)
\label{eq:gradient}
\end{equation}
\begin{equation}
    \mathrm{SNR}(t) = \frac{\mathbb{E}\left[\|\alpha_t x_0\|^2\right]}
       {\mathbb{E}\left[\|\sigma_t \epsilon\|^2\right]}= \frac{\alpha_t^2}{\sigma_t^2}\cdot
       \frac{\mathbb{E}\left[\|x_0\|^2\right]}
       {\mathbb{E}\left[\|\epsilon\|^2\right]}
\propto \frac{\alpha_t^2}{\sigma_t^2}
\label{equation:snr}
\end{equation}
\begin{equation}
    \mathbb{E}\left[\| \nabla_\theta\ell_t\|\right] \propto p(t) \cdot \mathrm{SNR}(t)=\frac{1}{T}\cdot\frac{\alpha_t^2}{\sigma_t^2}
    \label{equation:gradientsnr}
\end{equation}
The total learning signal accumulated during training can be viewed as the aggregation of gradient contributions from all denoising timesteps as in Eq. \ref{eq:total_contribution}. Since $\mathrm{SNR}(t)$ is a monotonically decreasing function of $t$, this sum is dominated by early timesteps with high SNR. Even though all timesteps are sampled equally often, their contributions to learning are highly imbalanced.
Consider timesteps in 2 regions $\mathcal{T}_{\text{high}}=\{t : \mathrm{SNR}(t) \ge \tau\},\quad
\mathcal{T}_{\text{low}}=\{t : \mathrm{SNR}(t) < \tau\}$, the total gradient becomes Eq. \ref{eq:totalgradient2regions}. Crucially, the contribution of each timestep to the total gradient is linear in $p(t)$. For any timestep t with non-zero $\mathrm{SNR}(t)$, $\sum_{t \in \mathcal{T}_{\text{low}}} \frac{1}{T}\uparrow$ while $\mathrm{SNR}(t)$ is unchanged, $\sum_{t \in \mathcal{T}_{\text{low}}}
\frac{1}{T}\,\mathrm{SNR}(t)\;\;\text{increases linearly with}\;\;\sum_{t \in \mathcal{T}_{\text{low}}} \frac{1}{T}$ then $\frac{\partial}{\partial p(t)}
\mathbb{E}\left[\|\nabla_\theta \mathcal{L}\|\right]=\mathrm{SNR}(t), \quad \forall t$. Therefore, although low-SNR timesteps are intrinsically harder, they are not gradient-silent. Their limited influence during standard training arises from insufficient exposure rather than a lack of informative signal. This observation motivates further investigation into how redistributing learning effort across timesteps affects model behavior, particularly in regimes. Rather than treating memorization as a binary or unintended artifact, this perspective suggests that memorization in diffusion models can be regulated through explicit control over where learning occurs along the denoising trajectory.

\begin{equation}
    \mathbb{E}\left[\|\nabla_\theta \mathcal{L}\|\right]
\propto
\sum_{t=1}^{T} p(t)\,\mathrm{SNR}(t)=
\sum_{t=1}^{T}
\frac{1}{T}\,\frac{\alpha_t^2}{\sigma_t^2}
\label{eq:total_contribution}
\end{equation}
\begin{equation}
    \mathbb{E}\left[\|\nabla_\theta \mathcal{L}\|\right]
\propto \sum_{t \in \mathcal{T}_{\text{high}}} \frac{1}{T}\,\mathrm{SNR}(t) + \sum_{t \in \mathcal{T}_{\text{low}}} \frac{1}{T}\,\mathrm{SNR}(t)
\label{eq:totalgradient2regions}
\end{equation}

\section{Memorization control}
\subsection{Normal timestep sampling and gradient control}
Instead of sampling diffusion timesteps uniformly, we propose to sample timesteps from a normal distribution $t \sim \mathcal{N}(\mu, \sigma^2), t\in [0, T]$. Under the denoising objective, the effective contribution of a timestep $t$ to learning can be approximated as Eq. \ref{eq:contribution}. With normal sampling, the probability density function $p(t)$ as Eq. \ref{eq:normaldist}. As shown in Eq. \ref{eq:derivcontribution}, the impact of the distribution parameters on learning is analyzed through the derivative of the contribution $C(t)$ with respect to the mean $\mu$. 
\begin{equation}
    C(t)=p(t)\cdot\mathrm{SNR}(t)
    \label{eq:contribution}
\end{equation}
\begin{equation}
    p(t)=\frac{1}{\sigma \sqrt{2\pi}}\mathrm{exp}(-\frac{(t-\mu)^2}{2\sigma^2})
    \label{eq:normaldist}
\end{equation}
\begin{equation}
    \frac{\partial C(t)}{\partial \mu} = \mathrm{SNR}(t)\cdot \frac{t-\mu}{\sigma^2} p(t)
\label{eq:derivcontribution}
\end{equation}
Increasing $\mu$ amplifies the contribution of timesteps $t>\mu$ while suppressing the contribution of timesteps $t<\mu$. Therefore, $\mu$ provides a direct mechanism to shift learning emphasis toward later timesteps, which typically correspond to low-SNR regions of the diffusion process. Besides, $\sigma$ controls the breadth of this emphasis, determining how sharply learning is concentrated around the mean.
\subsection{Confidence-interval based parameterization}
Although the normal distribution offers flexible control over timestep sampling, $\mu$ and $\sigma$ are unintuitive on the discrete diffusion time axis. To address this, we introduce a confidence-interval-based parameterization that directly specifies the region of timesteps where learning is emphasized. Specifically, we define a confidence interval $[c_l, c_h]$ such that $\mathbb{P}(c_l\leq t \leq c_h)=\gamma$ where $\gamma \in (0,1)$ denotes a predefined coverage ratio. Under the Gaussian assumption, the corresponding mean $\mu$ and standard deviation $\sigma$ are given by Eq. \ref{eq:mustd} where $z_\gamma$ is the quantile of the standard normal distribution associated with coverage $\gamma$.
\begin{equation}
    \mu=\frac{c_l+c_h}{2}, \quad \sigma=\frac{c_h-c_l}{2z_\gamma}
    \label{eq:mustd}
\end{equation}
This confidence interval serves as a design parameter that explicitly encodes the learning focus along the diffusion trajectory. By adjusting the bounds $c_l$ and $c_h$, we can directly control which timesteps receive increased learning signal, without reasoning about abstract distributional parameters. 

\subsection{Tail-mass truncation and redistribution}
\label{sec:tail_mass}
\begin{figure}[t]
    \centering
    \begin{minipage}[c]{0.49\columnwidth}
        \centering
        \includegraphics[width=\linewidth]{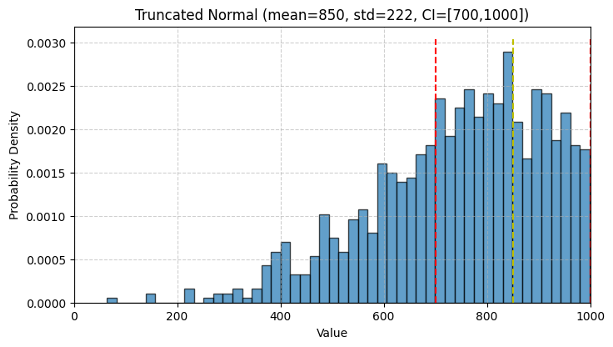}
    \end{minipage}
    \hfill
    \begin{minipage}[c]{0.49\columnwidth}
        \centering
        \includegraphics[width=\linewidth]{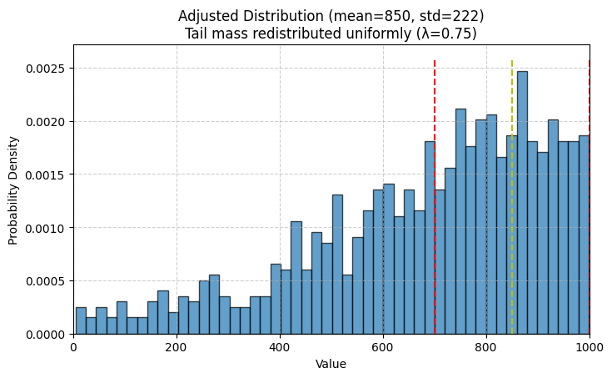}
    \end{minipage}

    \caption{Illustration of the truncated normal timestep sampling. Probability mass outside the valid timestep range is treated as tail mass and redistributed to ensure full coverage of all timesteps.}
    \label{fig:sec4.3}
\end{figure}

A normal distribution assigns non-zero probability mass outside the valid timestep range $[0, T]$. Let the tail probability be defined as Eq. \ref{eq:tail_mass} where $p(t)$ denotes the Gaussian timestep density. Na\"ively truncating the distribution to $[0, T]$ would discard this probability mass, potentially resulting in insufficient coverage of certain timesteps. As shown in Eq. \ref{eq:mixture_sampling}, to preserve full coverage of the diffusion horizon, we construct a mixture sampling distribution that combines a truncated normal distribution with uniform sampling where $p_{\mathrm{trunc}}(t)$ denotes the normal distribution truncated to the interval $[0, T]$. The mixing coefficient $\lambda$ is chosen to preserve the total probability mass of the original distribution. This mixture construction ensures that all timesteps retain non-zero sampling probability, while maintaining the intended learning bias induced by the confidence-interval-based normal distribution as in Fig. \ref{fig:sec4.3}.

\begin{equation}
P_{\mathrm{tail}}
=
\int_{-\infty}^{0} p(t)\,dt
+
\int_{T}^{\infty} p(t)\,dt,
\label{eq:tail_mass}
\end{equation}
\begin{equation}
\lambda
=
\frac{\int_{0}^{T} p(t)\,dt}
     {\int_{0}^{T} p(t)\,dt + P_{\mathrm{tail}}}.
\label{eq:lambda_def}
\end{equation}
\begin{equation}
p(t)
=
\lambda\, p_{\mathrm{trunc}}(t)
+
(1-\lambda)\frac{1}{T},
\label{eq:mixture_sampling}
\end{equation}
\section{Experiments and evaluation}
\begin{figure*}[ht]
    \centering
    \includegraphics[width=\textwidth]{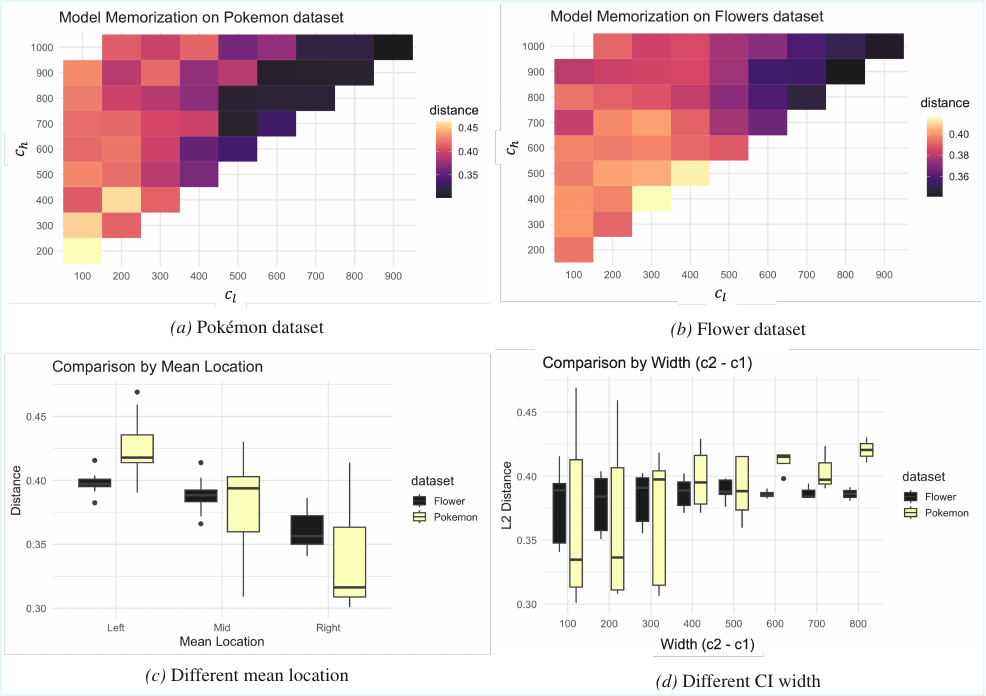}
    \caption{Distance of generated data and training data on (a) Pokémon and (b) Flower dataset with different $c_l$ and $c_h$ pairs (considering each diagonal) and the evaluation of model performance of different (c) mean location (1000 steps are divided into 3 categories - left, mid, and right) and (d) CI width on both datasets}
    \label{fig:comparison2dmodels}
\end{figure*}
\begin{figure*}[ht]
    \centering
    \includegraphics[width=\textwidth]{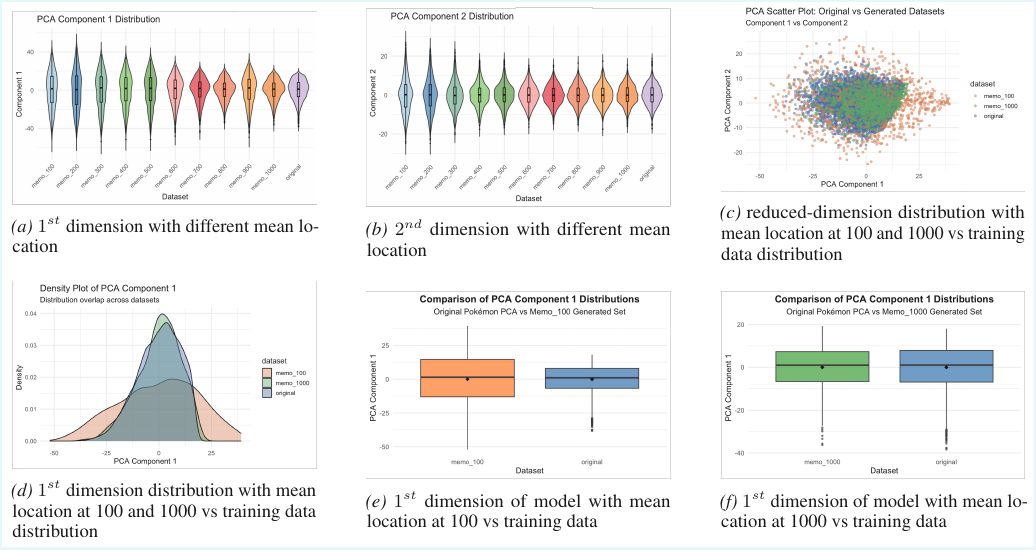}
    \caption{Difference of generated and training data on models with different mean location observed on 2 PCA components(a) and (b). Distribution of mean location at 100 and 1000 are selected to compare visually directly with the original (c, d, e, f) with fix CI=[700, 1000]}
    \label{fig:pcacomparison}
\end{figure*}
\subsection*{Experimental details}
To validate our hypothesis, we conduct experiments on two image datasets, Pokémon~\cite{djilax_pkmn_image_dataset} and Flowers-102~\cite{huggan_flowers102}, as well as one 1D time-series dataset, ECG5000~\cite{ECG5000}. Due to the differing data modalities, the image and 1D signal tasks employ distinct model architectures and preprocessing pipelines. However, in all cases, models are initialized from pretrained checkpoints on the same corresponding dataset to ensure a shared base level of generative knowledge before applying our proposed memorization control strategy. For 1D signal generation, we employ a dedicated diffusion architecture tailored to low-dimensional time-series data, directly modeling the original 140-dimensional signal without using a variational autoencoder to avoid unnecessary information compression. The 1D model follows a hierarchical U-Net design with increasing channel capacity across resolution levels and integrated attention at intermediate stages. This model is trained using with learning rate $1e-3$ and on 50 epochs to ensure sufficient baseline knowledge. 

Subsequently, memorization control is performed under a unified training setup across all datasets, where models are further trained for 30 epochs with a learning rate of $1\times10^{-4}$ using our proposed confidence-interval--based timestep sampling strategy. The input resolution of image and 1D signal are 64 and 140, respectively. Specifically, we define a confidence interval $[c_l, c_h]$ over the timestep range $[1,1000]$, with distribution parameters derived from a fixed Z-score $z=0.67449$, corresponding to a central $50\%$ confidence interval. We evaluate all valid CI pairs selected from $\{100, 200, \dots, 1000\}$, forming a triangular grid that systematically shifts training emphasis toward later timesteps.

\subsection*{Evaluation method}
To assess the impact of our proposals, we perform systematic evaluation on both image and 1D signal generation tasks. For the image datasets (Pokémon~\cite{djilax_pkmn_image_dataset} and Flowers-102~\cite{huggan_flowers102}), we generate 1,000 synthetic images from each trained model. We first compute the L2 distance between each generated image and the full training set to evaluate reconstruction fidelity then find the mean over training images. To assess distributional similarity more comprehensively, we reduce the dimensionality of both generated and training samples using principal component analysis (PCA) and then compare the distributions using Wasserstein distance~\cite{villani2008optimal} and Jensen-Shannon (JS) distance~\cite{lin1991jsdist}. These metrics collectively capture both how closely generated samples reproduce the training data and how well their overall distribution aligns with the real data distribution. For the 1D time-series dataset (ECG5000~\cite{ECG5000}), we generate 1,000 synthetic signals and compare each generated sample to the entire training set using the JS distance. This allows us to quantify how faithfully the generated signals follow the underlying distribution of the training data. 

\subsection{Image generation}
\begin{wrapfigure}{l}{0.5\textwidth}
    \centering
    \includegraphics[width=0.5\textwidth]{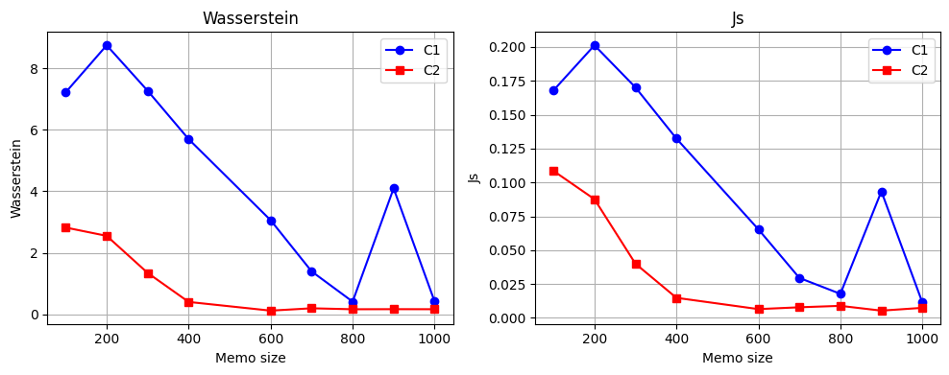}
    \includegraphics[width=0.5\textwidth]{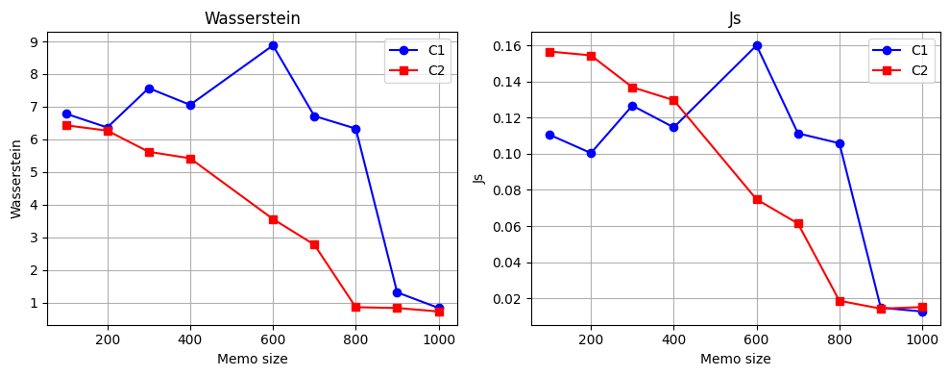}
    \caption{Wasserstein and JS distance of generated and training data distribution on Pokemon (first row) and Flower (second row) data (C1 and C2 are 2 dimension of reduced dimension data)}
    \label{fig:1ddistance}
    \vspace{-1.5cm}
\end{wrapfigure}
Fig. \ref{fig:comparison2dmodels} presents the evaluation of image generation models under different CI ranges on two datasets. As shown in sub-figures (a) and (b), for both datasets, our method is validated by observing each diagonal: the distance clearly decreases as the CI shifts to the right, which represents the generalization region. In addition, these sub-figures show that increasing the difference between $c_l$ and $c_h$ (making the distribution closer to a uniform distribution) leads to a larger distance between the generated data and the training data. To further study this effect, we apply hypothesis testing to examine whether the mean location and the CI width (gaps between $c_l$ and $c_h$) are correlated with model memorization.

This analysis is illustrated in Fig. \ref{fig:comparison2dmodels} (c) and (d). As the mean location moves to the right, the distance decreases significantly for both datasets. Hypothesis testing shows a strong and significant negative correlation between mean location and distance, with a correlation of -73.88\%. The 95\% confidence interval of the correlation ranges from -0.8212 to -0.6263. Regarding the effect of the gap between $c_l$ and $c_h$, Fig. \ref{fig:comparison2dmodels} (d) shows that a larger gap, which makes the distribution closer to uniform, results in a higher distance than a smaller gap. For the Pokémon dataset, when the width is 800, the minimum and maximum distances are both above 0.4. In contrast, when the width is 100, the distance distribution has a lower quartile (Q1) of 0.3 and an upper quartile (Q3) of 0.4. Hypothesis testing indicates a moderate positive correlation of 31.6\%, and this relationship is statistically significant, indicating a systematic effect rather than noise.

Additionally, we compare the PCA distributions of the generated data and the training data. As shown in Fig. \ref{fig:pcacomparison} (a) and (b), the distribution of the generated data becomes closer to the training data distribution as the mean of the CI increases. When directly comparing models with mean values of 100 and 1000, the model with a mean of 1000 produces a distribution that is much closer to the training data distribution. Additionally, this trend is consistent with the distance metrics shown in Fig. \ref{fig:pcacomparison}. For the first dimension C1, as the mean of the CI (memory size) increases from 100 to 1000, the Wasserstein distance decreases from 7.21 to 0.42, and the JS distance decreases from 0.168 to 0.011. This represents a reduction of approximately 94\% in Wasserstein distance and 93\% in JS divergence, indicating a much closer match between generated and training data distributions. Although there is a temporary increase at memory size 900, the overall downward trend remains clear. For the second dimension, the reduction is even more stable. The Wasserstein distance decreases from 2.82 at memory size 100 to around 0.17 at memory size 1000, while the JS divergence drops from 0.109 to 0.007. This corresponds to reductions of about 94\% and 93\%, respectively. After memory size 400, both metrics remain consistently low, showing that the generated data closely follows the training data distribution.

A similar pattern is observed on the Flower dataset. The Wasserstein distance decreases from 6.79 to 0.84 as the memory size increases from 100 to 1000, while the JS divergence drops sharply from 0.111 to 0.013 on the first dimension. Notably, both metrics show a significant reduction after memory size 800, where the Wasserstein distance falls from 6.33 to 1.31 and then to 0.84. For Flower Component 2, the Wasserstein distance decreases steadily from 6.43 to 0.73, and the JS divergence decreases from 0.157 to 0.015 as the memory size increases. This corresponds to an approximate 89\% reduction in Wasserstein distance and a 90\% reduction in JS divergence.

All numerical results illustrate that increasing the CI mean leads to consistent and substantial reductions in Wasserstein distance and JS divergence. They support our hypothesis that larger CI means produce generated data distributions that are significantly closer to the training data distributions.

\subsection{1D signal generation}
\begin{wrapfigure}{l}{0.5\textwidth}
    \centering
    \includegraphics[width=\linewidth]{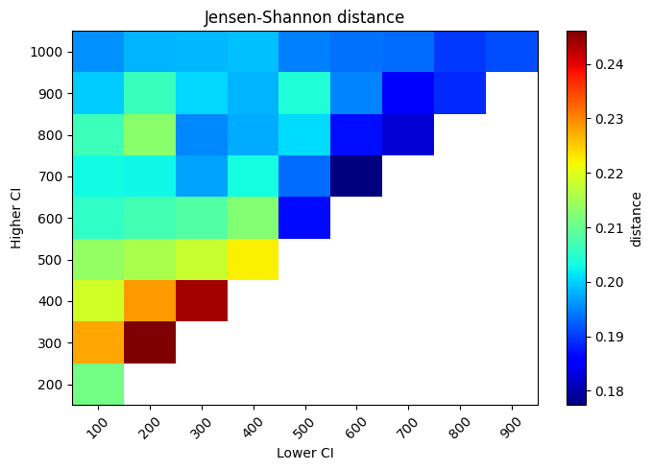}
    \caption{Distance of generated ECG5000 models and original training data}
    \label{fig:1djsdist}
    \vspace{0cm}
\end{wrapfigure}
\begin{figure*}[ht]
    \centering
    \includegraphics[width=\textwidth]{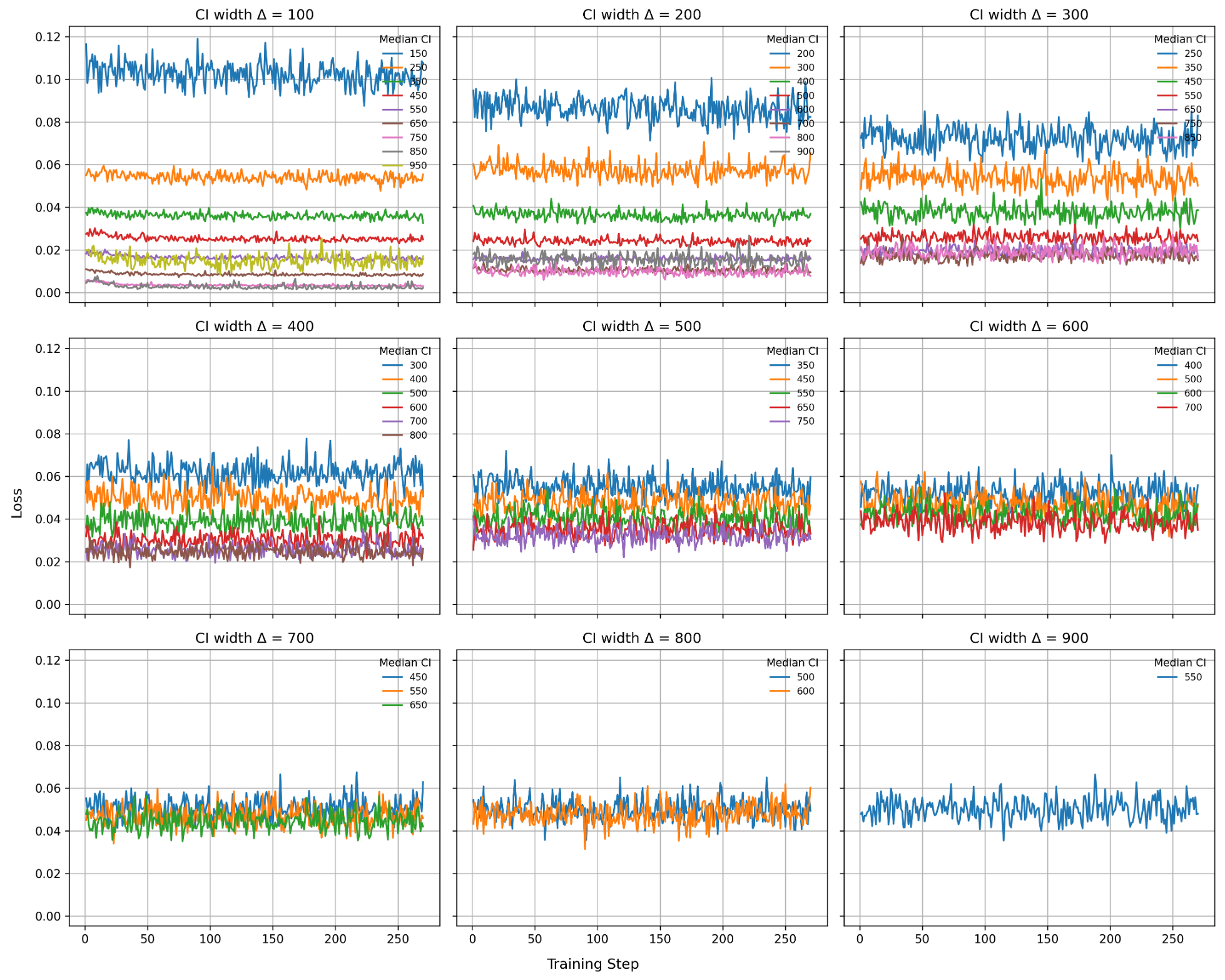}
    \caption{Training loss (last 30 epochs) comparison of models with different mean locations for each $c_l$ and $c_h$ gap}
    \label{fig:1dloss}
\end{figure*}

To show that our hypothesis is not limited to 2D data (i.e., images), we further validate it on a different task: 1D signal generation. As shown in Fig. \ref{fig:1djsdist}, the distance clearly decreases as the mean location moves toward the right. Moreover, Fig. \ref{fig:1dloss} compares the training loss in the last 30 epochs of models with different mean locations under various CI widths For a fixed CI width, models with larger mean locations consistently achieve lower loss values. This trend is clearly visible when comparing curves within each sub-figure: as the median CI increases, the loss shifts downward and remains more stable over training steps.
When the CI width is small (e.g., $\Delta = 100$ or $200$), the effect of mean location is strong. Models with smaller mean values show higher loss, while increasing the mean leads to a clear reduction in loss. For example, at $\Delta = 100$, increasing the median CI from 150 to 950 reduces the loss from approximately 0.10 to below 0.01. A similar pattern is observed for $\Delta = 200$ and $\Delta = 300$.
As the CI width increases, the loss values across different mean locations become closer. For moderate widths ($\Delta = 400$ to $600$), the loss curves start to overlap, and the difference caused by changing the mean location becomes smaller. When the CI width is very large ($\Delta \geq 700$), the loss values are nearly identical across different mean locations, indicating that the effect of the mean location is largely reduced.
This result is consistent with the observations from the image generation task and further supports our proposed hypothesis. Specifically, it confirms that the mean location of the CI plays a key role in controlling the distance between generated data and training data, regardless of the data dimensionality. The same trend observed in 2D image generation also appears in 1D signal generation, showing that the effect is not task-specific. This consistency across different data types demonstrates that our hypothesis captures a general and systematic behavior of the model, rather than an effect limited to a particular dataset or modality.

\section{Conclusion}
In this paper, we study memorization in diffusion models from a denoising-centric perspective. We show that, despite uniform timestep sampling during training, denoising steps contribute unequally to learning due to differences in signal-to-noise ratio, which leads to biased learning dynamics and can induce memorization. Based on this analysis, we propose a confidence-interval–based timestep sampling strategy that explicitly controls where learning is emphasized along the denoising trajectory. By parameterizing timestep sampling using intuitive lower and upper bounds, our method directly specifies the region of denoising steps that receive increased learning signal. To ensure full coverage of the diffusion horizon, we further introduce a tail truncation and redistribution scheme that preserves probability mass outside the valid timestep range while maintaining the intended learning bias. This design provides an intuitive and effective mechanism to shift model focus between memorization and generalization regimes.

Through extensive experiments on both image generation and 1D signal generation tasks, we empirically validate our hypothesis. Our results consistently show that the mean location of the confidence interval plays a dominant role in controlling memorization behavior, while the CI width governs the strength of this effect. The observed trends are robust across datasets, data dimensionalities, and evaluation metrics, demonstrating that the proposed approach captures a general and systematic property of diffusion models rather than a task-specific phenomenon.
{

}

\begin{thebibliography}{10}

\bibitem{ho2020denoising}
Jonathan Ho, Ajay Jain, and Pieter Abbeel.
\newblock Denoising diffusion probabilistic models.
\newblock {\em Advances in neural information processing systems}, 33:6840--6851, 2020.

\bibitem{11268708}
Thanhthuy Luyen, Thuy~Phuong Vu, Congduc Truong, Xuanque Nguyen, Minhhuy Le, and Quang~Vuong Pham.
\newblock Genndt: Conditional 1d nondestructive testing signal generation framework with diffusion model.
\newblock In {\em 2025 International Conference on Advanced Technologies for Communications (ATC)}, pages 1--6, 2025.

\bibitem{halder2024memorization}
Indranil Halder.
\newblock From memorization to generalization: a theoretical framework for diffusion-based generative models.
\newblock {\em arXiv e-prints}, pages arXiv--2411, 2024.

\bibitem{achilli2025memorization}
Beatrice Achilli, Luca Ambrogioni, Carlo Lucibello, Marc M{\'e}zard, and Enrico Ventura.
\newblock Memorization and generalization in generative diffusion under the manifold hypothesis.
\newblock {\em Journal of Statistical Mechanics: Theory and Experiment}, 2025(7):073401, 2025.

\bibitem{song2025selective}
Kiwhan Song, Jaeyeon Kim, Sitan Chen, Yilun Du, Sham Kakade, and Vincent Sitzmann.
\newblock Selective underfitting in diffusion models.
\newblock {\em arXiv preprint arXiv:2510.01378}, 2025.

\bibitem{10656527}
Chen Chen, Daochang Liu, and Chang Xu.
\newblock { Towards Memorization-Free Diffusion Models }.
\newblock In {\em 2024 IEEE/CVF Conference on Computer Vision and Pattern Recognition (CVPR)}, pages 8425--8434, Los Alamitos, CA, USA, June 2024. IEEE Computer Society.

\bibitem{bonnaire2025why}
Tony Bonnaire, Rapha{\"e}l Urfin, Giulio Biroli, and Marc Mezard.
\newblock Why diffusion models don{\textquoteright}t memorize: The role of implicit dynamical regularization in training.
\newblock In {\em The Thirty-ninth Annual Conference on Neural Information Processing Systems}, 2025.

\bibitem{yoon2023diffusion}
TaeHo Yoon, Joo~Young Choi, Sehyun Kwon, and Ernest~K Ryu.
\newblock Diffusion probabilistic models generalize when they fail to memorize.
\newblock In {\em ICML 2023 workshop on structured probabilistic inference $\{$$\backslash$\&$\}$ generative modeling}, 2023.

\bibitem{gu2023memorization}
Xiangming Gu, Chao Du, Tianyu Pang, Chongxuan Li, Min Lin, and Ye~Wang.
\newblock On memorization in diffusion models.
\newblock {\em arXiv preprint arXiv:2310.02664}, 2023.

\bibitem{buchanan2025edgememo}
Sam Buchanan, Druv Pai, Yi~Ma, and Valentin~De Bortoli.
\newblock On the edge of memorization in diffusion models, 2025.

\bibitem{ye2025provable}
Zeqi Ye, Qijie Zhu, Molei Tao, and Minshuo Chen.
\newblock Provable separations between memorization and generalization in diffusion models.
\newblock {\em arXiv preprint arXiv:2511.03202}, 2025.

\bibitem{kim2025diffusion}
Juyeop Kim, Songkuk Kim, and Jong-Seok Lee.
\newblock How diffusion models memorize.
\newblock {\em arXiv preprint arXiv:2509.25705}, 2025.

\bibitem{djilax_pkmn_image_dataset}
{djilax}.
\newblock Pokémon image dataset.
\newblock Kaggle: \url{https://www.kaggle.com/datasets/djilax/pkmn-image-dataset}, 2021.

\bibitem{huggan_flowers102}
{HugGAN Community}.
\newblock huggan/flowers-102-categories: Flower dataset with 102 categories.
\newblock \url{https://huggingface.co/datasets/huggan/flowers-102-categories}, 2022.
\newblock Dataset accessed via Hugging Face.

\bibitem{ECG5000}
Yanping Chen and Eamonn Keogh.
\newblock Ecg5000 time series classification dataset, 2018.

\bibitem{villani2008optimal}
C{\'e}dric Villani et~al.
\newblock {\em Optimal transport: old and new}, volume 338.
\newblock Springer, 2008.

\bibitem{lin1991jsdist}
J.~Lin.
\newblock Divergence measures based on the shannon entropy.
\newblock {\em IEEE Transactions on Information Theory}, 37(1):145--151, 1991.

\end{thebibliography}
\end{document}